\PassOptionsToPackage{table}{xcolor}
\documentclass[sigconf,authorversion=true]{acmart}

\AtBeginDocument{%
  \providecommand\BibTeX{{%
    \normalfont B\kern-0.5em{\scshape i\kern-0.25em b}\kern-0.8em\TeX}}}

\copyrightyear{2023} 
\acmYear{2023} 
\setcopyright{acmlicensed}
\acmConference[GECCO '23]{Genetic and Evolutionary Computation Conference}{July 15--19, 2023}{Lisbon, Portugal}
\acmBooktitle{Genetic and Evolutionary Computation Conference (GECCO '23), July 15--19, 2023, Lisbon, Portugal}
\acmPrice{15.00}
\acmDOI{10.1145/3583131.3590487}
\acmISBN{979-8-4007-0119-1/23/07}

\usepackage[capitalize]{cleveref}
\usepackage{tabularx}
\usepackage{amsfonts}

\renewcommand{\epsilon}{\varepsilon}

\newcommand{\D}{\mathcal{D}}
\renewcommand{\G}{\mathcal{G}}

\usepackage[table]{xcolor}
\newcommand{\colorrows}{\rowcolors{2}{white}{cyan!15}}

\newcommand{\repo}{\url{https://github.com/cavalab/fomo-gecco23}}
\begin{document}

\title{Optimizing fairness tradeoffs in machine learning with multiobjective meta-models}

\author{William G. {La Cava}}
\email{william.lacava@childrens.harvard.edu}
\orcid{0000-0002-1332-2960}
\authornotemark[1]
\affiliation{%
  \institution{Boston Children's Hospital}
  \city{Boston}
  \state{MA}
  \country{USA}
  \postcode{02215}
}


\begin{abstract}
  Improving the fairness of machine learning models is a nuanced task that requires decision makers to reason about multiple, conflicting criteria. 
The majority of fair machine learning methods transform the error-fairness trade-off into a single objective problem with a parameter controlling the relative importance of error versus fairness.
We propose instead to directly optimize the error-fairness tradeoff by using multi-objective optimization. 
We present a flexible framework for defining the fair machine learning task as a weighted classification problem with multiple cost functions.  
This framework is agnostic to the underlying prediction model as well as the metrics. 
We use multiobjective optimization to define the sample weights used in model training for a given machine learner, and adapt the weights to optimize multiple metrics of fairness and accuracy across a set of tasks. 
To reduce the number of optimized parameters, and to constrain their complexity with respect to population subgroups, we propose a novel meta-model approach that learns to map protected attributes to sample weights, rather than optimizing those weights directly. 
On a set of real-world problems, this approach outperforms current state-of-the-art methods by finding solution sets with preferable error/fairness trade-offs.
\end{abstract}



\maketitle


\section{Introduction}

Machine learning (ML)-based risk prediction models assist decision making in several high-risk fields, from criminal sentencing~\cite{berkFairnessCriminalJustice2018} to lending~\cite{nobleAlgorithmsOppression2018} to health care~\cite{ashanaEquitablyAllocatingResources2021}.  
Systematic unfairness in these models can and does lead to an outsized negative impact on marginalized subpopulations~\cite{obermeyerDissectingRacialBias2019}. 
In recourse, fair ML methods have been developed to mitigate biases in performance among subpopulations~\cite{barocasFairnessMachineLearning2019}. 
Typically these methods work by defining a measure of fairness appropriate to the domain, and then casting this metric as an additional loss or constraint to be satisfied during model development. 

Unfortunately, constraining machine learning (ML) models to be fair can come with serious negative side effects~\cite{kleinbergInherentTradeoffsFair2016}. 
In the risk prediction / classification context, there is a well-known error-fairness tradeoff for measures such as equal false positive (FP) / false negative (FN) rates and low overall error~\cite{kearnsPreventingFairnessGerrymandering2018}. 
In short, constraining a model to distribute ``mistakes" equally among subgroups typically requires the model to perform sub-optimally with respect to the overall population. 
Furthermore, when subgroups in the data have different base rates of the outcome/label, a fairness-fairness tradeoff arises: one cannot ensure equal false positive (FP) and false negative rates without miscalibrating the model for one or more subgroups~\cite{pleissFairnessCalibration2017}. 
Thus, model designers and decision makers must be extremely careful in picking fairness constraints for their problems, as they form strict trade-offs with error as well as other notions of fairness. 

Previous authors have noted that, due to the inherent trade-offs that exist, it is important to consider solution sets rather than individual models~\cite{hardtEqualityOpportunitySupervised2016,kearnsEmpiricalStudyRich2018,quadriantoRecyclingPrivilegedLearning2017}. 
In practice, model developers do not know beforehand what type of trade-off is possible; e.g., for a given dataset, it may be possible to limit the error rate differences between groups to 1\% by allowing a 1\% larger overall error rate; or, limiting the error rate differences to 1\% could require the use of a model with 50\% higher overall error. 
Because one doesn't know \emph{a priori} what the trade-off ought to be, we advocate here for returning the \emph{Pareto frontier} of models that represent best trade-offs of the objectives, rather than specifying a fixed cutoff. 
However, few works quantitatively compare the solution sets generated by fair ML methods as part of their empirical validation (\citep{lacavaGeneticProgrammingApproaches2020} being a notable exception).

With these goals in mind, we propose and study an evolutionary multi-objective optimization (EMO) framework referred to as fairness-oriented multiobjective optimization (FOMO).
At a high level, FOMO defines the fair machine learning task as one of solving a weighted classification problem with multiple objectives. 
In order to make this problem tractable for large datasets, we propose an approach in which we use EMO to learn the parameters of a meta-model that maps sensitive attributes of individuals (e.g. race, gender, income) to sample weights during ML training.   
In addition to reducing problem complexity, using a meta-model constrains the complexity of the mapping that is learned from groups to fairness costs - potentially acting as a regularizer between samples. 
Finally, we focus on comparing the relative performance of methods for their ability to generate high-quality solution \textit{sets}, rather than single point comparisons. 

In the following section, we describe the types of fairness and error tradeoffs that exist in binary classification contexts, and the methods for fair ML that have been proposed thus far.  
We then describe the FOMO method and what distinguishes it from previous work. 
In the experiments section, benchmark FOMO on a number of real-world classification tasks in comparison to related methods, and analyze the resultant solution sets between treatments. 
We conclude with a discussion of these results and promising new directions they suggest. 

\section{Background}


In the fairness setting for binary classification, each sample consists of a label, $Y \in \{0,1\}$, a set of $d$ features, $X \in \mathcal{X}$, and a set of $p$ \textit{protected attributes}, $X' \in \mathcal{X'}$. 
Note that $X'$ may or may not be a subset of $X$, depending on the constraints of the problem~\cite{thomasPreventingUndesirableBehavior2019}. 
Our training dataset consists of $N$ individuals/samples of these elements: 
$\mathcal{D} = \{(X,X',Y)_i\}_{i = 1}^{N}$.
The protected attributes are used to define population groups, which are subsets of samples that share a group identity (e.g., women over 65 years of age).   
These groupings form a collection of subsets, $\mathcal{G}$. 
We denote an individual's membership in a group $G \in \mathcal{G}$ by writing $X' \in G$. 
Let $R(X) \in [0,1]$ be the risk prediction model, and $\hat{Y} \in \{0,1\}$ be the binary classifier formed by applying a fixed threshold to $R(X)$. 

\subsection{Definitions of Fairness}

We now must define measures of fairness that guarantee, in some specific sense, that $R$ does not discriminate with respect to $\G$~\cite{hardtEqualityOpportunitySupervised2016}. 
Although dozens of measures of fairness exist in ML literature~\cite{castelnovoZooFairnessMetrics2021}, nearly all boil down to one of three basic notions of fairness: \emph{Independence}, \emph{Separation}, and \emph{Sufficiency}. 

\emph{Independence} holds that the risk scores should be independent of population groups. 
It can be expressed as
$$
\Pr[ R | X' \in G ]  = \Pr [R] 
\text{, for all } G \in \G. 
$$
In classification settings, independence is often referred to as \emph{demographic parity}. 
Under this notion, risk scores should be indistinguishable across demographic groups. 
Although independence is sometimes warranted~\cite{fouldsAreParityBasedNotions2020}, it can be undesirable in applications where demographics are associated with important risk factors for predicting the outcome.  
In such cases, independence can actually produce more unfairness~\cite{hardtEqualityOpportunitySupervised2016}.
For this reason we do not consider measures of independence in this work.

\emph{Separation} holds that the risk scores should be independent of population groups \emph{given the outcome}. 
It can be expressed as
$$
\Pr[ R | Y, X' \in G ]  = \Pr [R | Y] 
\text{, for all } G \in \G. 
$$
This notion is equivalent to requiring equality in false positive and false negative rates across groups, a criteria also known as \textit{Equalized Odds} \cite{hardtEqualityOpportunitySupervised2016}.  

\emph{Sufficiency} holds that the outcome should be independent of population groups \emph{conditioned on the risk scores}. 
It can be expressed as
$$
\Pr[ Y | R, X' \in G ]  = \Pr [Y | R] 
\text{, for all } G \in \G. 
$$
This notion underlies fair calibration measures such as multicalibration~\cite{hebert-johnsonMulticalibrationCalibrationComputationallyIdentifiable}.
\emph{Calibration} measures the extent to which risk scores actually represent the probability of an event occurring. 
In other words, among individuals with a risk score $r$, approximately an $r$ fraction should be positively labelled. 
A model is then fairly calibrated if $ \Pr[ Y | R=r, X' \in G ]  = r$, for all $r \in R$ and $G \in \G$. 

\subsection{Inherent Tradeoffs}

It is impossible to satisfy these three notions of fairness simultaneously~\cite{barocasFairnessMachineLearning2019}, and in all but trivial cases, separation and sufficiency cannot be jointly satisfied either. 
Although separation and sufficiency are almost always desirable, they are not jointly \emph{achievable} when there are differences in outcome rates among groups, unless the classifier is perfect. 
Even under relaxed conditions, separation can only be satisfied by worsening the performance of the model on some groups~\cite{pleissFairnessCalibration2017}.

\begin{figure}
    \includegraphics[width=\columnwidth]{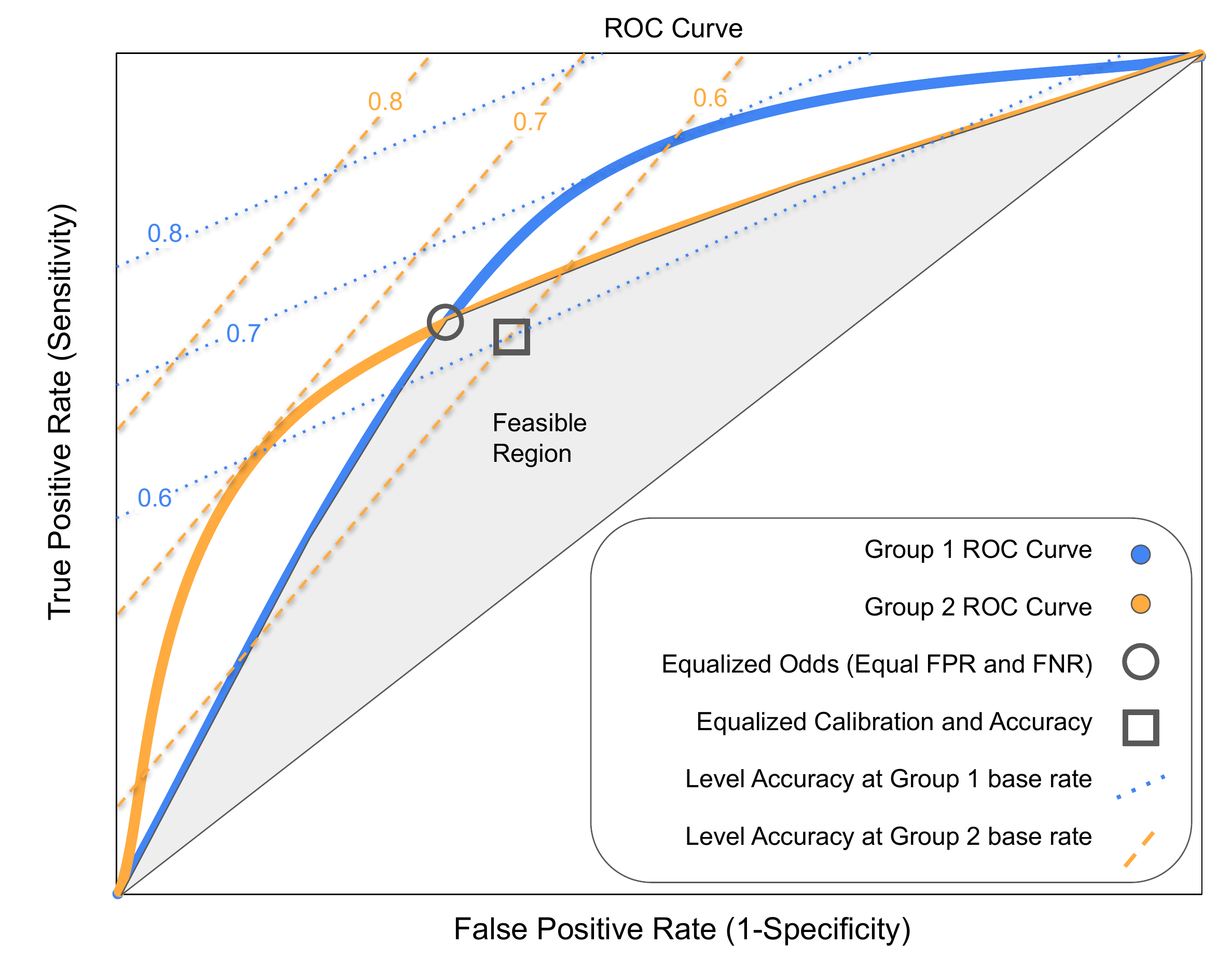}
    \caption{
        A visual guide to fairness tradeoffs in classification of two groups. 
        Numbers are purely illustrative. 
    }
    \label{fig:roc}
\end{figure}

We illustrate these tradeoffs using a receiver operator characteristic (ROC) curve in \cref{fig:roc}, which plots the true positive rate of a risk prediction model against its false positive rate, for different classification thresholds. 
The curve essentially defines the ``best we can do" in terms of these two classification metrics as a function of the classification threshold.   
In this example, we assume the two groups that have different base rates of the outcome, i.e. 
$\Pr[Y|X' \in G_1] \neq \Pr[Y|X' \in G_2]$. 

The performance of the model is plotted for two groups, $G_1$ (blue) and $G_2$ (dark yellow). 
The feasible region (gray) denotes potential operating points for which the model could perform equally for both groups. 
However, the points in the feasible region that do not lie on an ROC curve require making the classifier worse in terms of false positives, false negatives, or both, for one or both groups.   
The circle at the intersection of the two curves represents the single operating point for which the false positive and false negative rates of the model are equal for both groups (equalized odds), while still being at an ``optimal'' operating point for both groups. 
In other words, all fair operating points besides the circle require a strict trade-off with the model's FP and/or FN rates.  

Because the base rates differ between groups, an additional problem arises: operating at the circle means that the model is miscalibrated for one or both of the groups. 
Consider the dotted lines, which represent levels of accuracy for the model on each group. 
Accuracy is the number of correct predictions divided by the total number of predictions, and is tightly linked to calibration.  
Namely, the \textit{slope} of the lines of equal accuracy is determined by the base rates of the outcome within the group~\cite{pleissFairnessCalibration2017}. 
Therefore, the model is both calibrated and equally accurate for both groups wherever the lines of equal accuracy intersect within the feasible region. 
We denote such an intersection by the square, which, notably, is not on either group's ROC curve.
The square and circle will only intersect, i.e. equalized odds and fair calibration be simultaneously achieved, if the two two groups have identical base rates or the classifier is perfect on both groups. 
Thus we have not only an error-fairness, but also a fairness-fairness tradeoff among operating points in the model.

We can relax equalized odds by dividing it into simpler constraints that only ask for equivalence among the positive samples negative ones.  
We define the (soft) false positives (FP) for $R$ on a specific group as 
\begin{equation}
    FP(R,G) = \sum_i^N [ R_i | Y_i = 0, X'_i \in G].
\end{equation}
We can express group-wise false negatives analogously, as
\begin{equation}
    FN(R,G) = \sum_i^N [ 1-R_i | Y_i = 1, X'_i \in G].
\end{equation}
We use $FP(R)$ and $FN(R)$ to denote the FN/FP rates on the entire dataset.

These relaxations are useful for two reasons. 
First, a weaker constraint is easier to mix with other desirable notions of fairness and model performance. 
Second, the practical implication of a false positive prediction and false negative prediction can differ drastically by task.  
For example, falsely recommending a patient to be discharged from a hospital (false negative) may cause more harm than falsely recommending them to be admitted (false positive), since a false discharge puts patients at risk of an adverse health event without access to care. 

In this work we use two definitions of fairness proposed by \citet{kearnsPreventingFairnessGerrymandering2018} to capture differences in false positives and false negatives among multiple, potentially overlapping subgroups in the data. 
Together these measures are referred to as Subgroup Fairness and are defined below. 

\begin{definition}[False Positive Subgroup Fairness]
    \label{def:sfp}
    Fix a dataset $\mathcal{D}$ and risk model, $R(X)$. 
    Let $$\alpha_{FP}(G) = \Pr [ X' \in G, Y=0]$$ be the probability of the negative class within group $G$. 
    Define  
    $$
        \beta(R,G) = |FP(R) - FP(R,G)| 
    $$
    to denote the absolute difference in FP between the group and the whole population.  
    Then the \emph{False Positive Subgroup Fairness} (SFP) of $R$ on $\D$ is given by
    \[
        L_{SFP}(\D, R) = \max_{G \in \G}{ \alpha_{FP}(G) \beta(R,G)}
        . 
    \]
\end{definition}

\begin{definition}[False Negative Subgroup Fairness]
    \label{def:sfn}
    Fix a dataset $\mathcal{D}$ and risk model, $R(X)$. 
    Let $$\alpha_{FN}(G) = \Pr [ X' \in G, Y=1]$$ be the probability of the negative class within group $G$. 
    Define  
    $$
        \beta(R,G) = |FN(R) - FN(R,G)| 
    $$
    to denote the absolute difference in FN between the group and the whole population.  
    Then the \emph{False Negative Subgroup Fairness} (SFN) of $R$ is given by
    \[
        L_{SFN}(\D, R) = \max_{G \in \G}{ \alpha_{FN}(G) \beta(R,G)}
        . 
    \]
\end{definition}

Subgroup fairness captures the maximum deviation of a model's performance among any one group in $\G$, normalized by the probability of observing an individual from that group in the positive or negative labels. 
This normalization term is introduced to guarantee that the fairness measures will generalize to unseen samples. 

\subsection{Current Approaches to Fair Machine Learning}

\begin{table*}[tb]
    \centering
    \caption{The properties of a sample of approaches to fair ML. }
    \label{tbl:algs}
\colorrows
\begin{tabular}{l l l l l}
Fairness Treatment  &   Processing Stage   &   Fairness Metric     &   Grouping   &   ML Model
\\
\toprule
Reweighing~\cite{kamiranDataPreprocessingTechniques2012}  
    &   Preprocessing   
    &   Demographic Parity 
    &   Binary 
    &   Any 
    \\
Fair Feature Selection~\cite{rehmanFairFeatureSubset2022}    
    &   Preprocessing 
    &   Demographic Parity   
    &   Binary  
    &   Any 
    \\ \midrule 
Adversarial Debiasing~\cite{zhangMitigatingUnwantedBiases2018}  
    &   Training   
    &   Equalized Odds  
    &   Flexible
    &   Neural Network 
    \\
Differential Fairness~\cite{keyaEquitableAllocationHealthcare2021}    
    &   Training 
    &   Demographic Parity   
    &   Flexible
    & Neural Network 
    \\ 
FLEX~\cite{lacavaGeneticProgrammingApproaches2020}    
    &   Training 
    &  Subgroup Fairness   
    &  Flexible  
    & Symbolic Regression
    \\ 
Exponentiated Gradients~\cite{agarwalReductionsApproachFair2018}   
    &   Training   
    &   Equalized Odds, Demographic Parity
    &   Single attribute  
    &   Any
    \\
GerryFair~\cite{kearnsPreventingFairnessGerrymandering2018}   
    &   Training   
    &   Subgroup Fairness    
    &   Flexible  
    &   Any
    \\
\textbf{FOMO}    
    &   \textbf{Training} 
    &  \textbf{Any}    
    &  \textbf{Flexible}  
    & \textbf{Any}   
    \\ 
\midrule
Calibrated Equalized Odds~\cite{pleissFairnessCalibration2017}   
    &   Postprocessing 
    &  Relaxed Equalized Odds   
    &  Binary  
    & Any
    \\ 
MultiCalibration~\cite{hebert-johnsonMulticalibrationCalibrationComputationallyIdentifiable2018}   
    &   Postprocessing 
    &  Calibration   
    &  Flexible  
    & Any 
    \\ 
MultiAccuracy~\cite{pleissFairnessCalibration2017}   
    & Post 
    & Accuracy   
    & Flexible  
    & Any 
    \\ 
\bottomrule
\end{tabular}

\end{table*}

Many proposals for ensuring fair ML have been proposed and studied; we refer interested readers to the book by \citet{barocasFairnessMachineLearning2019} and reviews of \citet{chouldechovaFrontiersFairnessMachine2018} and \citet{castelnovoZooFairnessMetrics2021}.
These approaches can be taxonomized in different ways; we present a non-exhaustive list in \cref{tbl:algs} and a brief overview here. 

Fair ML approaches can be described by the stage of the ML pipeline on which they operate: preprocessing, i.e. before model training; in-processing or during training; and post-processing, which is after training has finished. 
Fair ML methods for pre-processing and post-processing are desirable because they are model agnostic; i.e., they only require access to the data before training or the fitted model after training, respectively. 
However, this flexibility means that the success of pre- and post-processing methods in ensuring fairness depends on success in the training phase. 
For example, even if data is ``de-biased" using reweighing~\cite{kamiranDataPreprocessingTechniques2012} or fair feature selection~\cite{rehmanFairFeatureSubset2022}, ML methods themselves can introduce new biases, e.g. choice of cost function or internal feature selection~\cite{thomasPreventingUndesirableBehavior2019,dworkFairnessAwareness2012}.
Post-processing methods like multiaccuracy~\cite{hebert-johnsonCalibrationComputationallyIdentifiableMasses2018} and equalized odds post-processing~\cite{hardtEqualityOpportunitySupervised2016} are similarly limited by the accuracy and fairness of the fitted model. 

There have been many proposals to adapt the training processes of specific algorithms in order to improve fairness.  
Methods like adversarial debiasing~\cite{zhangMitigatingUnwantedBiases2018} and differential fairness~\cite{fouldsIntersectionalDefinitionFairness2019} are examples of adaptations to neural network (NN) training. 
Two notable examples of adaptations to specific algorithms utilize EMO. 
\citet{quadriantoRecyclingPrivilegedLearning2017} suggested EMO as a way to train support vector machines with additional fairness constraints. 
\citet{lacavaGeneticProgrammingApproaches2020} used NSGA2~\cite{debFastElitistNondominated2000} with a fairness objective to improve the fairness of an existing symbolic regression method.
In both cases, the proposals were not model-agnostic, unlike the method we describe here. 

Many proposals for fair ML suggest adding fairness as an additional term to the loss function~\cite{berkConvexFrameworkFair2017,fouldsIntersectionalDefinitionFairness2019,dworkFairnessAwareness2012,kamishimaFairnessawareLearningRegularization2011}. 
Although these methods are, in principle, quite general, they require intervention on an ML method's underlying cost function, which make them cumbersome out of the box. 
As a result their implementations are typically demonstrated for single ML methods.  

Two notable examples of model-agnostic fair ML methods use cost-sensitive classification with out-of-the-box classifiers~\cite{agarwalReductionsApproachFair2018,kearnsPreventingFairnessGerrymandering2018}. 
Cost-sensitive classification is a variant of weighted classification in which each sample $(X,X',Y)_i$ has costs associated predicting the positive or negative class.
In this setting, the problem of satisfying an addition fairness constraint boils down to choosing appropriate weights for each positive/negative sample, and feeding these into the cost function (i.e. loss function) of the base learning algorithm. 

The fair, cost-sensitive classification problem can be solved using game-theoretic algorithms. 
\citet{agarwalReductionsApproachFair2018} proposed to use an exponentiated gradient descent algorithm, available as part of the \textit{fairlearn} package~\cite{bird2020fairlearn}. 
Although model agnostic, the main limitation of this proposal is that the definitions of groups ($\mathcal{G}$) is limited to a single, non-overlapping attribute. 
\citet{kearnsPreventingFairnessGerrymandering2018} proposed an extension to handle multi-attribute, potentially overlapping groups with a method named GerryFair, available as part of the AI-Fairness 360 package~\cite{bellamyAIFairness3602018}. 
In both cases, optimization consists of multiple iterations of a two player game, in which one player represents the classifier and the other, a fairness auditor. 
The result of either strategy is an ensemble classification model consisting of the models of each round, which are used in aggregate to make downstream predictions. 

These two approaches to fair ML are desirable in that they work on a large range of ML methods and the algorithms are provably convergent. 
However, they have certain shortcomings. 
First, they results in an ensemble of the desired ML model, that, while still of the original model class, may be much more complex than a single instance would be. 
Second, both methods requires pre-specification of the relative weight of a the fairness loss to the error loss. 
Third, both methods must employ specific loss functions to ensure that the problem remains convex. 

Based on this prior work, we were motivated to develop a general-purpose, fair ML approach which we call fairness-oriented multiobjective optimization, or FOMO. 
To our knowledge, FOMO is the only fair ML method that 1) adapts a model's training process, 2) pairs with (nearly) any ML method, 3) handles complex group definitions, 4) can handle multiple, potentially non-differentiable fairness or error objectives, and 5) naturally handles the optimization of solutions sets. 

\section{Fairness-Oriented Multiobjective Optimization}

FOMO injects fairness into the training process by adapting the sample weights of an ML classifier using EMO. 
Nearly all commonly used ML methods support weighted classification, including linear/logistic regression, support vector machines, decision trees, random forest, gradient boosted trees (XGBoost), and neural networks, among others. 
Weighted classification refers to classification in which the underlying loss function to be minimized accepts a weight for each sample. 
For example, a weighted loss function for minimizing classification errors would look like

\begin{equation*}
    L(w, \D,\hat{Y}) = \sum_i^N{ w_i \mathbb{I}[\hat{Y} \neq Y_i] }
    .
\end{equation*}

\begin{figure*}
    \includegraphics[width=\textwidth]{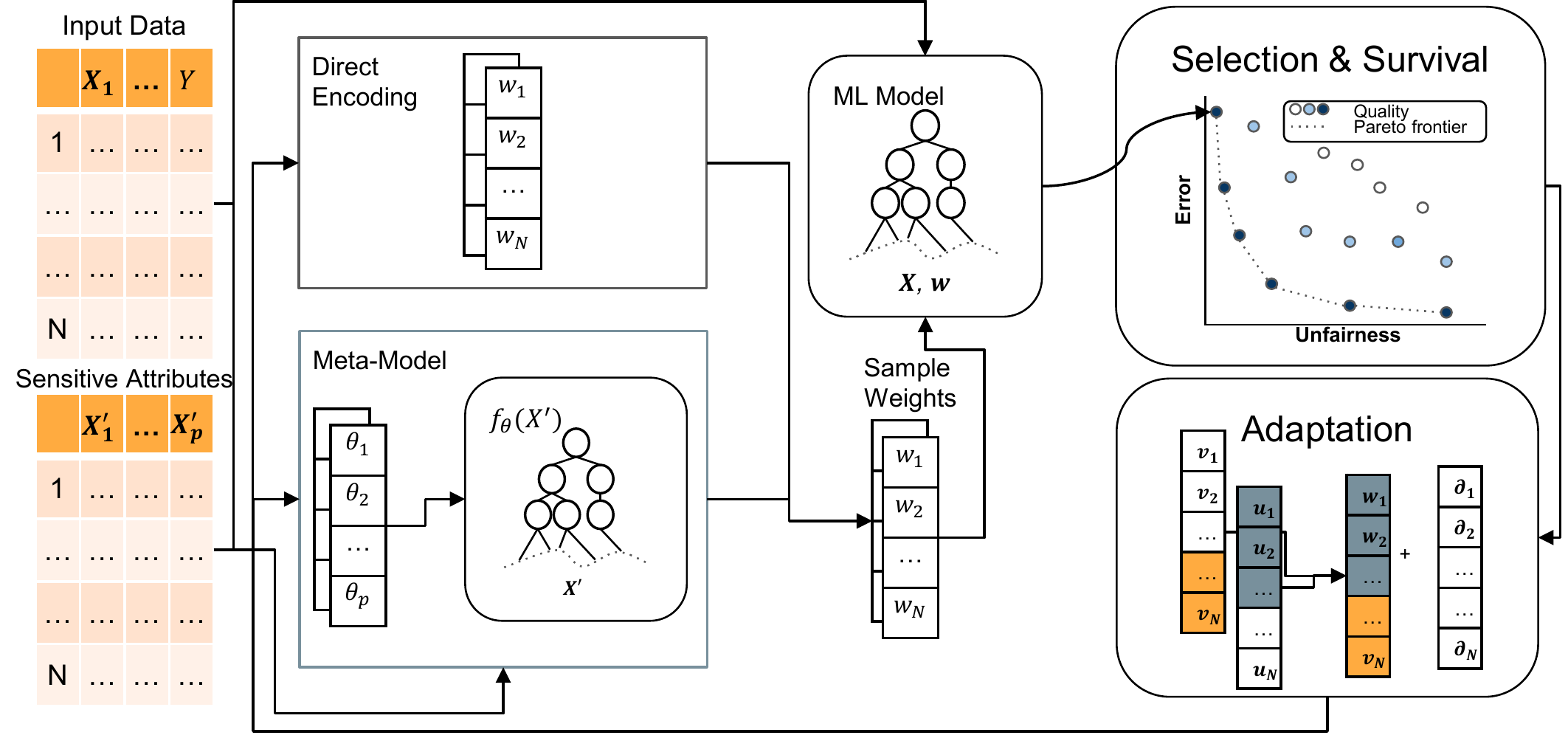}
    \caption{An overview of the multiobjective approaches tested.}
    \label{fig:diagram}
\end{figure*}

Given a set of accuracy and fairness objectives, our high-level goal is to generate a set of trained classifiers that represent high-quality, Pareto-efficient tradeoffs between the objectives. 
With FOMO, this is achieved by searching the space of sample weights, $w$, used to train those classifiers. 
A flow diagram of this approach is shown in \cref{fig:diagram}.

We test two variants of this approach. 
In the first, we directly optimize the weights associated with the cost function of the underlying ML model, using EMO to simultaneously satisfy multiple objectives. 
We refer to this as the ``direct" encoding as shown in \cref{fig:diagram}. 
It is defined formally as follow:

\begin{definition}[Direct Encoding Problem]
    \label{def:direct}
    Given a dataset $\D$ and a set of $M$ loss functions, solve the following:
\begin{align*}
    \min                &&   L_m(w, \D)             &&   m = 1,\dots,M   
    \\
    \text{such that}    &&   0 \leq w_i  \leq 1     &&   i=1,\dots,N
    \\
\end{align*}
\end{definition}

Typically one of the loss functions, $L_m$, corresponds to the loss function of the ML model, whereas the others may be chosen from existing fairness metrics (e.g. \cref{def:sfn,def:sfp}) that best align with the needs of the problem. 
However, one is free to define additional objectives and/or different accuracy objectives as well. 

To keep FOMO as generic as possible, it is implemented as an interface between EMO algorithms in  \textit{pymoo}~\cite{blankPymooMultiObjectiveOptimization2020} and ML methods in \textit{scikit-learn}~\cite{pedregosaScikitlearnMachineLearning2011a}. 
Therefore it supports multiple algorithms for EMO (NSGA2, NSGA3, MOEAD, etc) as well as any \textit{scikit-learn}-compatible classifier that supports sample weights during calls to its \textit{fit()} method. 
The package is available as free open source software: \repo.  

\subsection{Learning a meta-model of samples}

As sample sizes ($N$) increase, the direct encoding approach becomes intractable; 
one must learn a weight for each sample, for each model in the solution set. 
Instead, we propose to use optimize a surrogate model, $f_{\theta}(X'): \mathcal{X'} \rightarrow [0,1]^N$,  that estimates the sample weights as a function of the protected attributes (see \cref{fig:diagram}).  
EMO is then used to optimize the parameters of this model, $\theta$, re-formulating the problem as follows:

\begin{definition}[Meta-model Problem]
    \label{def:meta}
    Given a dataset $\D$, a set of $M$ loss functions, and a model class $f_{\theta}(X'): \mathcal{X'} \rightarrow [0,1]^N$, solve the following:
\begin{align*}
    \min    &&   L_m(f_{\theta}, \D, R)     &&   m = 1,\dots,M   
    \\
    \text{such that}    &&&&
    \\
    &&   0 \leq f_\theta(X'_i)  \leq 1  &&   i=1,\dots,N
    \\
    &&   0 \leq \theta_j  \leq 1  &&   j=1,\dots,|\theta|
    \\
\end{align*}
\end{definition}

The number of parameters to optimize, $|\theta|$, then becomes a function of the complexity of the model and the number of protected attributes, i.e. $|X'| = p$.  
In its simplest form, we treat $f_\theta$ as a linear model of $X'$, so that $|\theta| = p$. 
We also experiment with defining $f_\theta$ as a neural network (NN), for which  $|\theta|$ corresponds to the number of parameters in the network. 

The implications of using a meta-model extend beyond making FOMO more tractable. 
Whereas the direct encoding treats the relationship between model fairness and sample weights as a black box, using a meta-model encoding assumes the relationship to be a function of the protected attributes, better aligning with our fairness definitions. 
The complexity of the model class of $f_\theta$ then dictates how complex this relation is assumed to be. 
For example, using a linear $f_\theta$ assumes that there is a (multivariate) linear relationship between the protected attributes and model fairness. 
We hypothesize that this additional constraint on the complexity of the sample weights may lead to better generalization of the fairness measures by preventing overfitting to individual/small sets of samples.

\section{Experiments}

\begin{table*}
    \centering
    \small
    \caption{Properties of the datasets used for comparison.}
    \label{tbl:datasets}
    \colorrows
\begin{tabularx}{\textwidth}{lllrrrXX}
    Dataset     &   Source (link)  &   Outcome &   Samples &   Features    & Sensitive features  & Protection Types   & Number of groups ($|\mathcal{G}|$) \\ \hline
    Communities &   \href{http://archive.ics.uci.edu/ml/datasets/communities+and+crime}{UCI}    
                &   Crime rates     &   1994    &   122     &  18   &  race, ethnicity, nationality    &   1563    \\ 
    Adult       &   \href{https://archive.ics.uci.edu/ml/datasets/adult}{Census}    
                &   Income          &   2020    &   98      &   7  &   age, race, sex    &   78    \\
    Lawschool   &   \href{https://eric.ed.gov/?id=ED469370}{ERIC}    
                &   Bar passage     &   1823    &  17       &   4   &   race, income, age, gender   &   47 \\
    Student     &   \href{https://archive.ics.uci.edu/ml/datasets/student+performance}{Secondary Schools}    
                &   Achievement     &   395     &   43      &   5  &   sex, age, relationship status, alcohol consumption  &   22  \\
    \hline
\end{tabularx}    
\end{table*}

\begin{table}
    \centering
    \small
    \caption{Settings for the methods in the experiments.}
    \label{tbl:exp}
    \colorrows
\begin{tabularx}{\columnwidth}{llX}
    Treatment    &   ML Method   &   Error/Fairness Tradeoff ($\gamma$)  
    \\ \midrule
    GerryFair-LR
        & Logistic Regression
        &  100 values in $[0.001,1]$
        \\
    GerryFair-XGB
        & XGBoost~\cite{chenXGBoostScalableTree2016}
        &  
        \\
        - & -  &   Weight Encoding ($f_\theta$)
    \\ \midrule
    FOMO-LR
        & Logistic Regression
        & Direct
        \\
    FOMO-LR-Lin
        & Logistic Regression
        & Logistic Meta-Model 
        \\
    FOMO-LR-NN
        & Logistic Regression
        & NN Meta-Model 
        \\
    FOMO-XGB
        & XGBoost~\cite{chenXGBoostScalableTree2016}
        & Direct
        \\
    FOMO-XGB-Lin
        & XGBoost~\cite{chenXGBoostScalableTree2016}
        & Logistic Meta-Model 
        \\
    FOMO-XGB-NN
        & XGBoost~\cite{chenXGBoostScalableTree2016}
        & NN Meta-Model 
        \\

    \bottomrule
\end{tabularx} 
\end{table}

We test variants of FOMO and GerryFair on a set of four real-world classification tasks with known fairness concerns, detailed in \cref{tbl:datasets}. 
We compare eight different treatments of these methods that vary by the ML method and, for FOMO, the weight encoding. 
These methods are detailed in \cref{tbl:exp}. 
For each treatment, we test its ability to return a set of solutions that minimize subgroup fairness and maximize the area under the ROC curve (AUROC). 
We test two scenarios for subgroup fairness: minimization of the 1) subgroup FP and 2) FN rates (\cref{def:sfn,def:sfp}). 
For each treatment, dataset and scenario, we conduct 20 trials with 50/50 stratified train/test splits, and compare results on the test sets. 
To compare solution set quality, we use the well-known hypervolume measure~\cite{fonsecaImprovedDimensionsweepAlgorithm2006}, which allows the quality of a set of tradeoffs to be compared~\cite{chandEvolutionaryManyobjectiveOptimization2015}.
For two objectives, this corresponds to computing the area of the objective space that a solution set dominates with respect to a reference point. 
We use a reference point of $[1,1]$ and compute the hypervolume of the values of $(SF, 1-AUROC)$.  

To generate a set of solutions with GerryFair, we train models with 100 different values the error/fairness tradeoff parameter, $\gamma$ (see \cref{tbl:exp}). 
For FOMO, we set the population size to 100 and return the final set of solutions. 
For all treatments, we terminate training after 1 hour. 

Because FOMO is built on top of \textit{pymoo}~\cite{blankPymooMultiObjectiveOptimization2020} and \textit{scikit-learn}~\cite{pedregosaScikitlearnMachineLearning2011a}, there are number of options for the underlying EMO and ML methods. 
For simplicity, we use NSGA2~\cite{debFastElitistNondominated2000} for all runs and use on linear nad one non-linear ML classifiers: penalized logistic regression and gradient boosting~\cite{chenEthicalMachineLearning2020}.   
We use identical ML models in our comparisons to GerryFair. 
For the configuration of FOMO using a NN meta-model, the NN is defined as a multilayer perceptron with a single hidden layer of 10 neurons.

\section{Results}

\begin{figure*}
    \includegraphics[width=\textwidth]{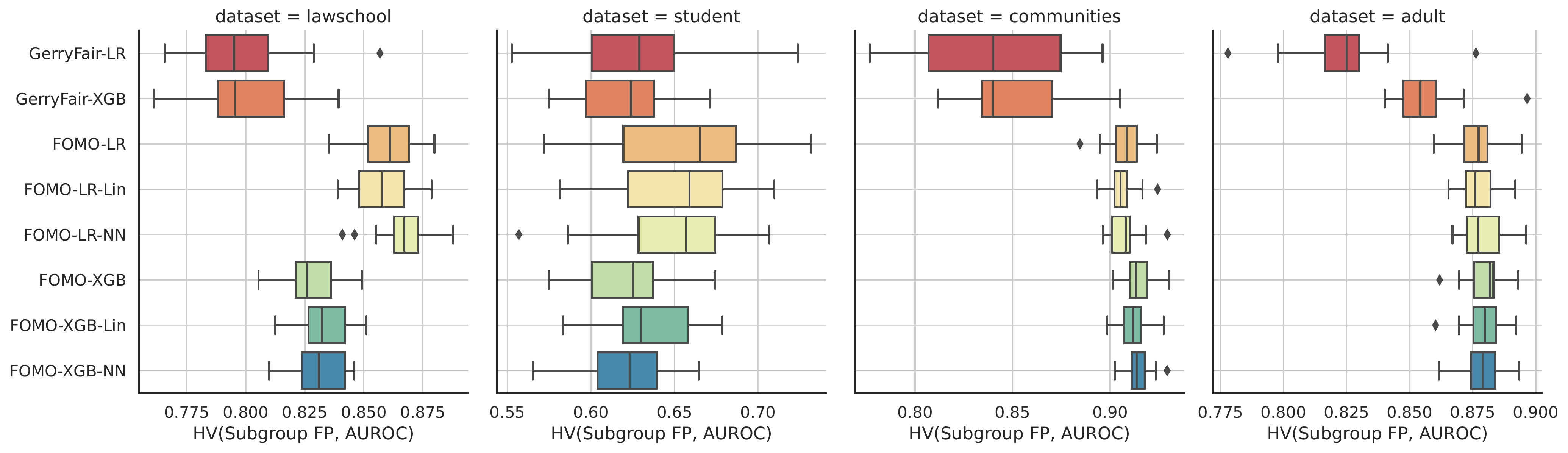}\\%
    \includegraphics[width=\textwidth]{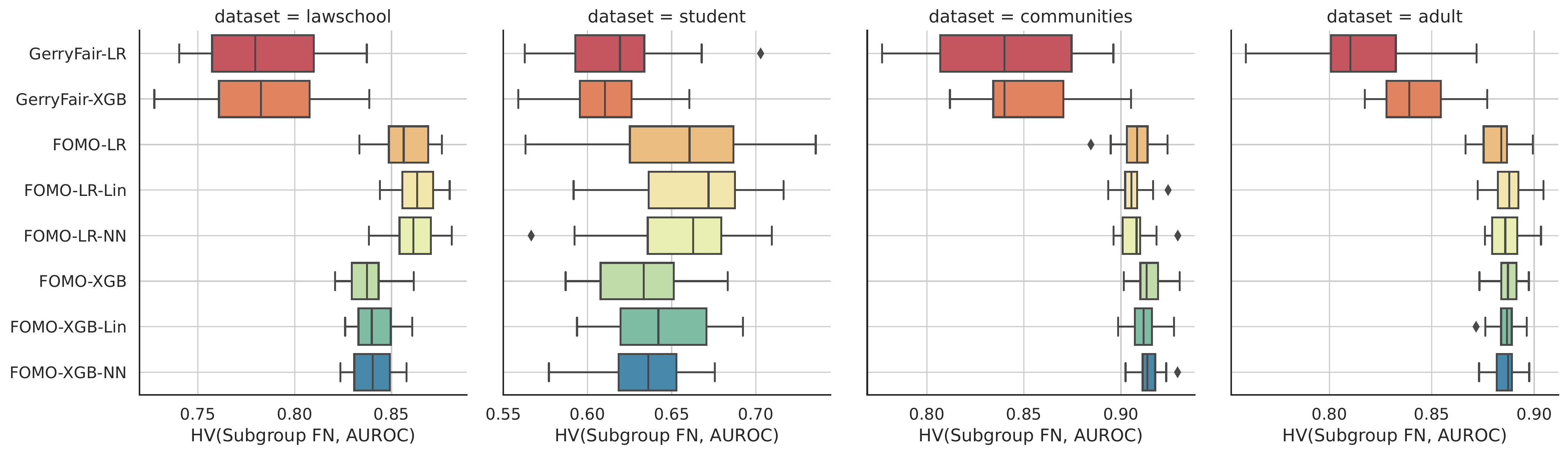}
    \caption{
        Hypervolume (HV) measure of error/fairness tradeoffs of the solution sets generated by each treatment (y-axis) across problems (columns).
        Top: optimizing for Subgroup FP fairness (\cref{def:sfp}) and AUROC. 
        Bottom: optimizing for Subgroup FN fairness (\cref{def:sfp}) and AUROC. 
        The box line denotes the median, the box width denotes the inter-quartile range (IQR), and whiskers extend to 1.5 times the IQR. 
        Points denote outliers. 
    }
    \label{fig:hv_boxes}
\end{figure*}

\begin{figure}
    \includegraphics[width=\columnwidth]{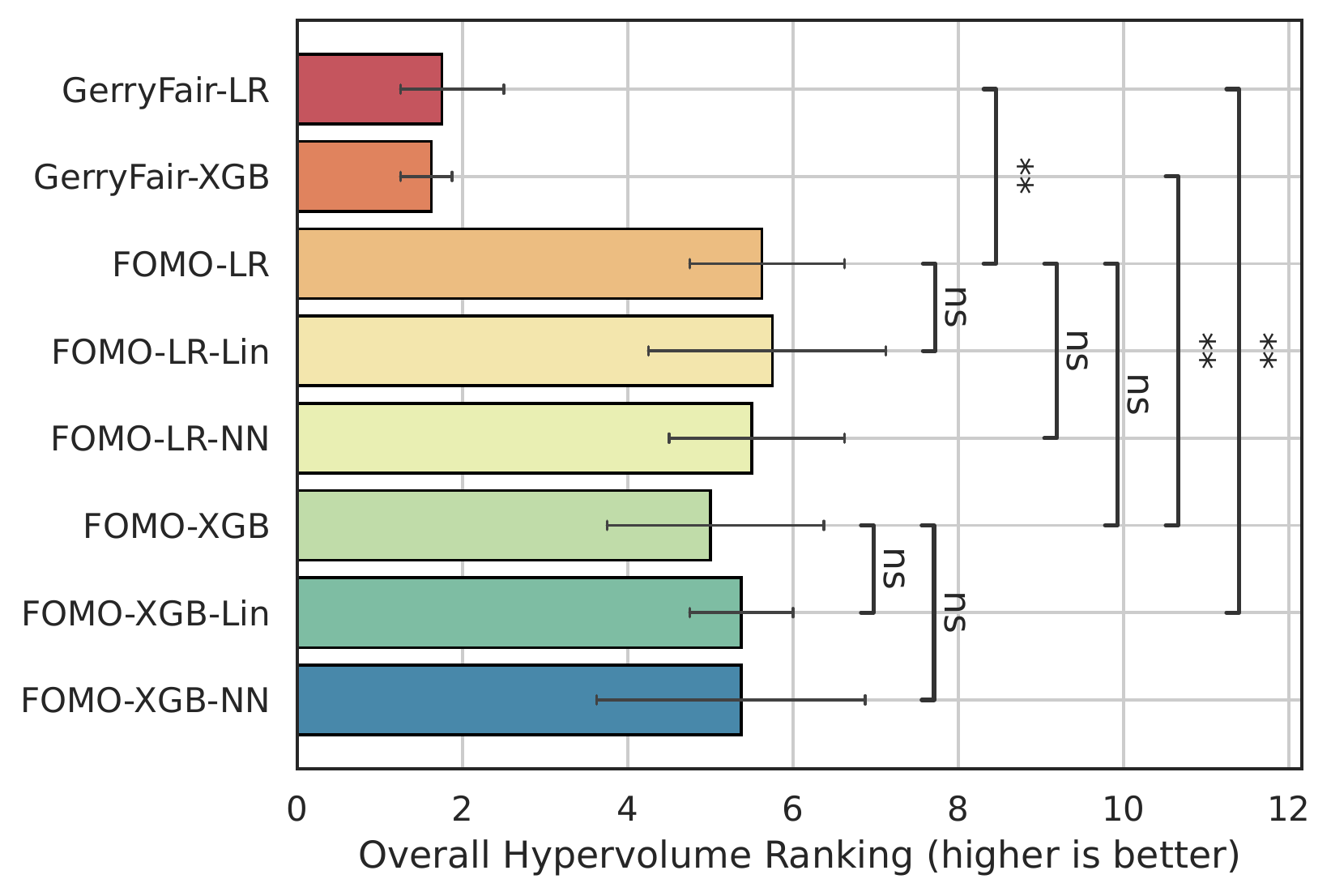}
    \vspace{1em}
    \caption{
        Average of the median rankings of treatments by the hypervolume metrics in \cref{fig:hv_boxes}, across problems. 
        Annotations denote the result of pairwise Mann-Whitney statistical tests with Bonferroni corrections. 
        ns: $p <= 1.00$;
        *: $0.01 < p <= 0.05$; 
       **: $0.001 < p <= 0.01$; 
       ***: $0.0001 < p <= 0.001$. 
    }
    \label{fig:ranks}
\end{figure}

The hypervolume comparisons between methods on each dataset are shown in \cref{fig:hv_boxes}. 
For each dataset and ML method, FOMO-based treatments return a larger median hypervolume measure than GerryFair-based treatments, with the exception of the student dataset for which only the logistic regression FOMO models outperform GerryFair. 
We compare overall hypervolume rankings in \cref{fig:ranks}.  
Across problems and fairness metrics, FOMO outperforms GerryFair for both logistic regression and gradient boosted models ($p \leq 9.270e-03$). 

We note that the use of a meta-model for encoding the weights does not significantly improve the hypervolumes of solution sets in FOMO on these problems. 
However, it does significantly reduce the number of optimized parameters, as shown in \cref{tbl:encoding}.

\begin{figure*}
    \includegraphics[width=\linewidth]{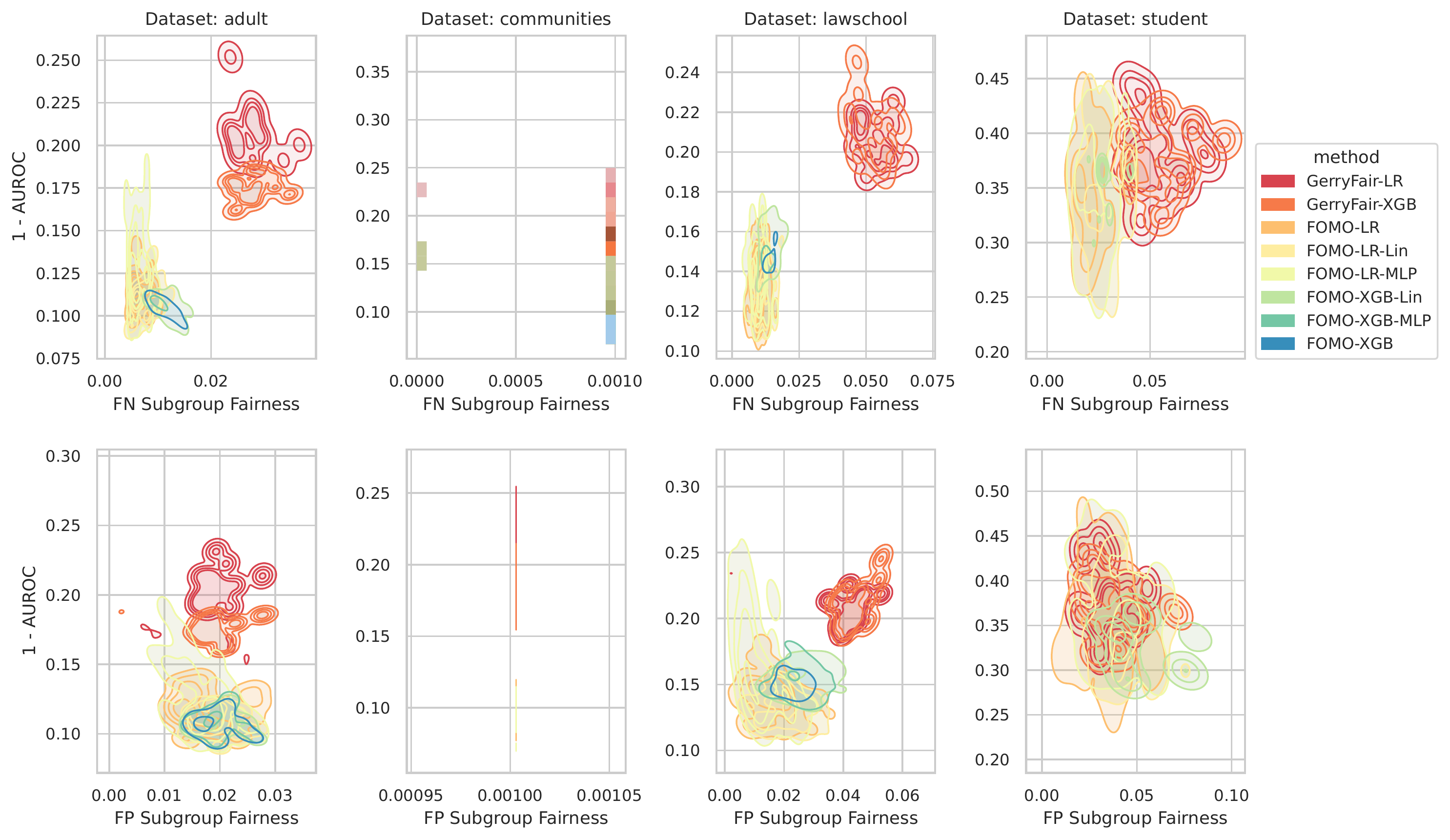}
    \caption{
        Distributions of the Pareto frontier of models returned by each method, evaluated on test sets. 
        Subplots from left to right are different datasets, and with top and bottom optimizing for FN and FP subgroup fairness, respectively.  
        The contours are generated using a kernel density estimate of the observed model distributions over all trials. 
        For the communities dataset, a bivariate histogram is shown instead.  
        }
    \label{fig:pareto}
\end{figure*}

In \cref{fig:pareto}, we visualize the distributions of solution sets generated by different methods for each problem. 
In general, we see that the solution set distributions generated by FOMO-based methods tend to dominate those generated by GerryFair. 
One exception is the FP Subgroup Fairness optimization on the student dataset, where the solution set distributions heavily overlap. 
On the communities datasets, methods tend to converge to a single value of subgroup fairness, for which the FOMO-based methods tend to have better AUROC performance. 
We also note that, although the solution sets returned by each method are of size 100, the set of Pareto optimal models decreases when the models are evaluated on the test sets. 

\begin{table}
    \centering
    \caption{Number of parameters optimized by FOMO (i.e. dimension of the design space) for different encoding strategies and datasets. }
    \label{tbl:encoding}
    \colorrows
    \begin{tabular}{llr}
        Dataset     &   Encoding    & Number of Parameters  
        \\
        \toprule
        Communities &&
        \\
            &   Direct              &   997
            \\
            &   Logistic Meta-model &   19 
            \\
            &   NN Meta-model       &   201
            \\
        Adult   &&  
        \\
            &   Direct              &   1010
            \\
            &   Logistic Meta-model &   8
            \\
            &   NN Meta-model       &   91 
            \\
        Lawschool   &&
        \\
            &   Direct              &   912
            \\
            &   Logistic Meta-model &   5
            \\
            &   NN Meta-model       &   131
            \\
        Student &&
        \\
            &   Direct              &   198
            \\
            &   Logistic Meta-model &   6
            \\
            &   NN Meta-model       &   71 
        \\
        \bottomrule
    \end{tabular}
\end{table}

\section{Discussion and Conclusions}

Ensuring that ML models behave fairly is a central concern in a world that is increasingly dependent on automated decision-making. 
However, improving fairness during training requires model developers to navigate several important, problem-dependent trade-offs. 
In this paper, we have illustrated the interplay of equalized odds, fair calibration, and model accuracy in the context of risk prediction and binary classification. 
We have proposed a method, FOMO, that uses EMO to solve a weighted classification problem in which the tradeoff space of multiple accuracy and fairness objectives is explicitly optimized. 
We introduced a novel encoding strategy for the weighted classification problem that reduces the number of parameters one needs to optimize, thereby allowing this approach to scale to large datasets. 
Through a set of experiments on real-world classification datasets with known fairness concerns, we have found that FOMO finds significantly better sets of trade-offs between fairness and accuracy objectives than comparable, model-agnostic fair ML approaches. 
We demonstrated this both for linear and non-linear ML models, as well as for controlling false positive and false negative rates across rich subgroups of the population.  
We hypothesized that using a meta-model approach to encoding the sample weights would additionally improve the ability of FOMO to find better accuracy/fairness trade-offs, but did not find this to be the case among the problems we studied. 

Although we find the results in general to be promising, limitations of these results should be noted. 
First, the experiment is currently limited to relatively small datasets (hundreds to thousands of samples). 
Although these datasets have been used in several related works~\cite{kearnsEmpiricalStudyRich2018,lacavaGeneticProgrammingApproaches2020}, additional evaluation on larger real-world datasets is needed, especially to increase dataset representation of minority subgroups.  
In addition, we limited our comparison to one existing fair ML approach - GerryFair - since it was the only other model-agnostic training method allowing flexible group definitions that we have found (see \cref{tbl:algs}). 
Whereas this allows for an apples-to-apples comparison, it means we cannot rule out that other fair ML approaches may find models exhibiting better trade-offs for these problems. 
Nonetheless, the results suggest that FOMO is a promising way to optimize the training process of a classifier to achieve desirable error/fairness trade-offs, across ML methods, tasks, and fairness metrics.

\section{Code Availability}
The FOMO software is available from \url{https://github.com/cavalab/fomo}. 
Code for the experiments is available from \repo.

\begin{acks}
  This work was partially supported by National Institutes of Health grant R00-LM012926 from the National Library of Medicine. 
\end{acks}

\bibliographystyle{ACM-Reference-Format}
\bibliography{Fairness}

\end{document}